\documentclass[10pt]{article} 
\usepackage[preprint]{tmlr}

\usepackage{amsmath,amsfonts,bm}

\def\eqref#1{equation~\ref{#1}}

\def\1{\bm{1}}

\DeclareMathAlphabet{\mathsfit}{\encodingdefault}{\sfdefault}{m}{sl}
\SetMathAlphabet{\mathsfit}{bold}{\encodingdefault}{\sfdefault}{bx}{n}

\usepackage{hyperref}
\usepackage{url}
\usepackage{natbib}
\usepackage{graphicx}
\usepackage{amsmath}
\usepackage{makecell}
\usepackage[utf8]{inputenc} 
\usepackage[T1]{fontenc}    
\usepackage{hyperref}       
\usepackage{url}            
\usepackage{booktabs}       
\usepackage{amsfonts}       
\usepackage{nicefrac}       

\usepackage{microtype}      
\usepackage{xcolor}         

\usepackage[most]{tcolorbox}
\usepackage{colortbl}
\usepackage{multirow}
\usepackage{graphicx}
\usepackage{lipsum}
\usepackage{tablefootnote}
\usepackage{tcolorbox}
\usepackage[tikz]{bclogo}
\usepackage{subfigure}
\usepackage{subcaption}
\usepackage{multirow}
\usepackage{wrapfig}

\title{Agents in the Large: Perception-Centered Architecture for Persistent Agents}

\author{\name Shihan Dou$^*$ \quad Haoxiang Jia$^*$ \quad Shichun Liu$^*$ \quad Feng Chen \quad Chenhao Huang \quad Yujiong Shen\\[1em] Shaofan Liu \quad Jiayi Chen \quad Jiahang Lin\quad Honglin Guo \quad Qianyu He\quad Minghao Guo \quad Ziyi Ye\\[1em] Pluto Zhou \quad Tao Gui \quad Qi Zhang \quad Xuanjing Huang
\\[1em] \addr NLP Lab, Fudan University \quad \quad PL Lab, Peking University\\[0.25em]
\addr CCDS, Nanyang Technological University \quad \quad Hunyuan Team, Tencent
}

\def\month{MM}  
\def\year{YYYY} 
\def\openreview{\url{https://openreview.net/forum?id=XXXX}} 

\begin{document}

\maketitle

\begingroup
\renewcommand\thefootnote{}
\footnotetext{$^{*}$Equal contribution. Each author may list their name first. Contact: \texttt{shihandou@foxmail.com}, \texttt{haoxiangjia@stu.pku.edu.cn}, \texttt{plutozhou096@foxmail.com}, and \texttt{tgui@fudan.edu.cn}.}
\endgroup

\begin{abstract}

Cognitive language agents have achieved substantial progress by equipping language models with memory, tools, and decision-making procedures, enabling agents to reason and act in interactive environments.
Existing frameworks largely cast these agents as systems for solving user-specified, bounded tasks.
An increasingly important goal is for language agents to provide persistent assistance in long-lived settings where user needs, context, and service procedures persist and change, and to remain useful across the broad range of tasks that arise over time.
Yet we still lack a framework to characterize persistent AI agents, organize existing work, and guide future development.
To this end, we propose a \textbf{Per}ception-Centered \textbf{A}rchitecture for \textbf{Per}sistent \textbf{A}gents (\textbf{Pera}). 
Pera describes a persistent agent organized around perception and control components that continually perceive service-relevant signals from episodic task executions, internal context, and changes in the surrounding environment, and use these signals to construct lifecycle tasks.
These tasks drive the ongoing operation and adaptation of the agent's service procedures.
We use Pera to retrospectively organize recent work, examine a detailed case study, and offer forward-looking insights for building more capable persistent agents.
Just as software engineering moved from programming in the small to programming in the large, Pera frames the evolution of language agents as an analogous architectural transition toward long-lived, adaptive intelligence systems.

\end{abstract}

\section{Introduction}

Software development once underwent a major transition from \emph{programming in the small} to \emph{programming in the large}~\citep{deremer1976programming}. 
As software came to support larger systems over longer periods of operation~\citep{lehman1980programs}, the focus of design expanded from correctly implementing individual programs~\citep{parnas1972criteria,parnas1971information} to sustaining systems that provide reliable service over time.
Software architecture~\citep{perry1992foundations,garlan1993introduction} came to address the organization of system components, interactions, and the constraints and rationale required to satisfy system-level requirements. In parallel, research on software evolution and dependability treated continuing change, maintenance, fault tolerance, recovery, and reliable service as lifecycle concerns~\citep{patterson2002recovery,garlan2004rainbow,de2013software,li2024generative}.
Thus, the object of design expanded from a bounded program execution to a persistent system that must remain reliable as requirements and operating conditions change~\citep{oreizy1998architecture}.

\begin{figure}[t]
\centering
\includegraphics[width=0.9\textwidth]{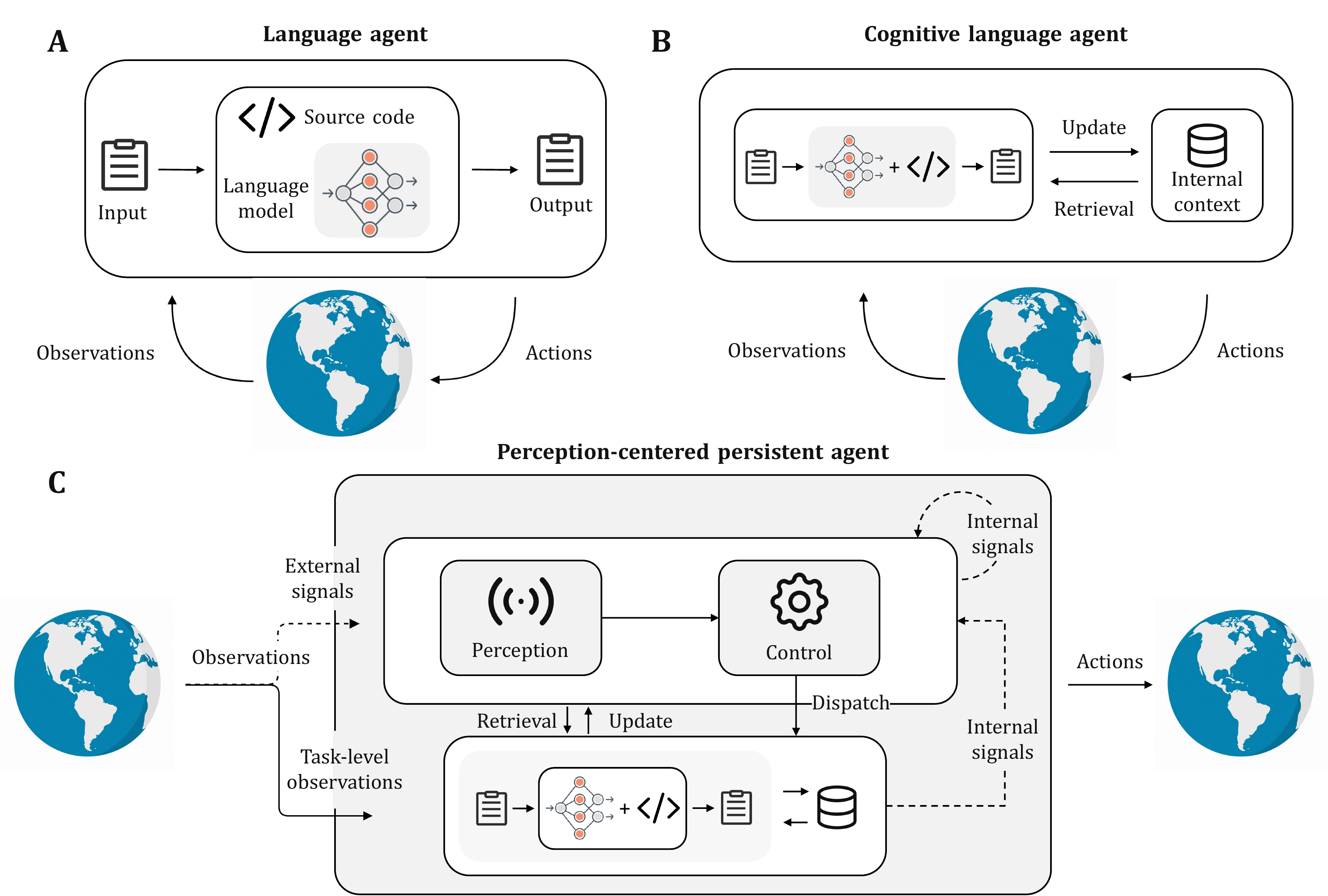}
\caption{
Comparison of three language agents. 
\textbf{A:} Language agents place a language model within an agent decision procedure implemented in source code, and interact with the external environment through grounding.
\textbf{B:} Cognitive language agents augment this interaction loop with internal context (i.e., memory) and cognitive actions such as reasoning, retrieval, and updating, allowing retained experience to inform subsequent tasks.
\textbf{C:} To sustain effective service in a long-lived setting, the agent must continually perceive changes in itself and its environment and proactively adapt accordingly.
Perception-centered persistent agents introduce active perception and lifecycle-level control to notice such changes and proactively update the agent and its service.
}
\label{fig:Pera-comparison}
\end{figure}

Language agents are now approaching an analogous architectural transition~\citep{fan2026cognitive}.
As their capabilities and operational horizons expand, they are increasingly expected to become persistent agents~\citep{fang2025comprehensive,gao2025survey,zheng2026lifelong} that provide persistent assistance~\citep{li2025hello} in long-lived settings~\citep{xu2026toward,zhu2026your}, beyond the completion of any single user-specified task.
However, most language agents are still architected as cognitive systems~\citep{sumers2023cognitive,wang2024survey,yao2022react} that coordinate language models, memory, tools, and decision-making procedures around a current user task~\citep{yao2022react,shinn2023reflexion,schick2023toolformer,yu2026agentic,du2026survey}.
Although recent agent systems have been run continuously for tens of hours while attempting a single complex task, empirical evaluations show that their effective task horizons remain much shorter and that progress often plateaus during extended runs~\citep{starace2025paperbench,kwa2026measuring,xu2026theagentcompany}.
This degree of autonomy still falls short of persistent agency.

Consider computer-use agents, one of the most actively studied classes of language agents.
Current evaluations of computer-use agents focus on completing open-ended tasks in real or simulated computer environments~\citep{sager2026comprehensive,wang2026computer}.
A more autonomous computer-use agent should do more than complete tasks explicitly assigned by the user~\citep{nushi2019guidelines,lu2025proactive,li2026proevent}.
It should continuously perceive the user's activities~\citep{yang2026contextagent,tang2026proagentbench} and changes across the computer environment~\citep{pasternak2025beyond,li2026proevent} and infer needs that have not yet been articulated~\citep{lu2025proactive,lan2026peap}.
It should also recognize when its current decision procedure~\citep{cox2005metacognition,hu2025automated,lin2026agentic} is inadequate for anticipated future tasks and revise it accordingly~\citep{zhang2024agent,du2026survey,zhang2026self}.
In other words, persistent assistance requires an agent not only to decide how to address the task at hand, but also to infer what additional work is needed and assess whether its current procedures remain adequate.

This example illustrates a broader transition from cognitive language agents, or \emph{agents in the small}, to persistent agents, or \emph{agents in the large}
(Figure~\ref{fig:Pera-comparison}).
The focus of agent design expands from solving a user-specified task to providing persistent assistance throughout the lifetime of a long-lived setting.
A user-specified task provides an explicit objective, but a long-lived setting presents signals whose relevance to future service must be actively recognized.
It also exposes far more information than a resource-bounded agent can process continuously.
Therefore, \emph{active perception} is key to this transition~\citep{bajcsy1988active,bajcsy2018revisiting,lu2025proactive}.
Active perception forms the interface through which an evolving, open-ended setting becomes observable to lifecycle-level control.
It allows the agent to anticipate emerging needs and prepare for changes without waiting for explicit instructions.
What is missing is an architectural account of persistent agents that specifies which components must exist beyond the cognitive core and how perceived change becomes work the agent performs on itself.

To this end, we propose \textbf{Pera}, the \textbf{Per}ception-Centered \textbf{A}rchitecture for \textbf{Per}sistent \textbf{A}gents, a conceptual framework that organizes the core components and operating principles of persistent agents.
Pera augments cognitive language agents with \emph{perception} and \emph{control} components that support persistent operation across the lifecycle of a setting. 
It distinguishes episodic tasks, which complete specific pieces of work, from lifecycle tasks, which sustain and improve the agent's continuing service of a setting.
The perception component continually senses and processes service-relevant signals from both external and internal sources. 
External signals arise from changes in the environment, while internal signals arise from task execution and the agent's internal context.
The control component interprets these signals and formulates lifecycle tasks.
Lifecycle tasks can inspect and update the agent's context, revise its decision procedures, or anticipate emerging user needs and prepare relevant support before they are articulated as explicit requests.

We further use Pera to retrospectively organize recent work and clarify how existing efforts contribute to persistent agents, and prospectively suggest actionable directions for building future persistent agents.

\textbf{Paper organization.}
The remainder of the paper is organized as follows. 
We first review a historical shift in software system design and early work that incorporated perception into agent architectures in Section~\ref{sec:background}.
We explain why perception is essential for agents serving long-lived settings in Section~\ref{sec:from-agent}. 
Then, Section~\ref{sec:Pera} introduces the Pera framework and uses it to organize representative prior work.
Section~\ref{sec:case_study} further illustrates Pera through a detailed case study, and Section~\ref{sec:actionable-insights} offers several actionable insights for advancing persistent agents.
Finally, Section~\ref{sec:dis} discusses related agent frameworks, and Section~\ref{sec:conclusion} concludes the present paper.

\section{Background}
\label{sec:background}

Three architectural ideas provide the basis for our account of persistent agents. Programming in the small and programming in the large distinguish computation within individual units from control over relations among them~\citep{deremer1976programming}. Cognitive architectures for language agents organize the machinery for executing bounded tasks~\citep{sumers2023cognitive}. Active perception shows that information acquisition is itself guided by an agent's activities and information needs~\citep{hayes1995architecture}. Together, these ideas allow us to ask what must change architecturally when an agent's service persists beyond any single task.

\subsection{Programming in the small and programming in the large}

A software module can correctly implement its local function while the larger system remains incorrect because system properties can depend on relations among modules~\citep{ockerbloom1995architectural,garlan1995architectural}. Individually correct modules can rely on incompatible assumptions, expose mismatched interfaces, or interact in ways that violate constraints enforced by none of them. Such properties cannot be addressed solely through computation internal to an individual module. 

Programming in the small concerns computation within an individual program or module, whereas programming in the large concerns the organization of relations among such units~\citep{deremer1976programming}. When a property depends on interfaces, dependencies, or interactions that span multiple units, the architecture must explicitly organize those relations. Therefore, programming in the large does not make the computation inside each unit larger; it adds control over properties that no individual unit organizes. These two levels are complementary: programming in the large organizes relations among units implemented in the small without replacing their internal computation. Architectural scale is determined by the scope of activity and state placed under control, rather than by the amount of computation performed within a unit. We use this criterion to distinguish architectural scales in language agents.

\subsection{Language agents as agents in the small}

Cognitive architectures for language agents place a language model within a system of memory, actions, and decision-making procedures that together support task execution~\citep{sumers2023cognitive,fan2026cognitive}. During execution, observations update the task state, while the decision procedure uses reasoning, retrieval, and interaction with the environment to select subsequent actions. In task-oriented language agents, this machinery is typically organized around a bounded objective, that directs execution, determines which information is relevant, and defines when the task is complete. We call tasks organized around such an objective an \emph{episodic task}, where \emph{episodic} refers to the scope of the tasks rather than a type of memory. 

We call an architecture an \emph{agent in the small} when the episodic task is the highest-level unit that it explicitly organizes. An episodic task may involve extensive planning, tool use, interaction, testing, and revision, and could extend across many actions or hours of execution~\citep{ma2024agentboard,starace2025paperbench,kwa2026measuring}. The agent could also retain long-term memory, learn reusable procedures, and reuse experience across tasks. Neither extended execution nor persistent state enlarges the architectural scope so long as they remain subordinate to the active episodic objective. Architectural scope expands only when persistent conditions and relations across tasks themselves become objects of control.

\begin{figure}
    \centering
    \includegraphics[width=0.9\linewidth]{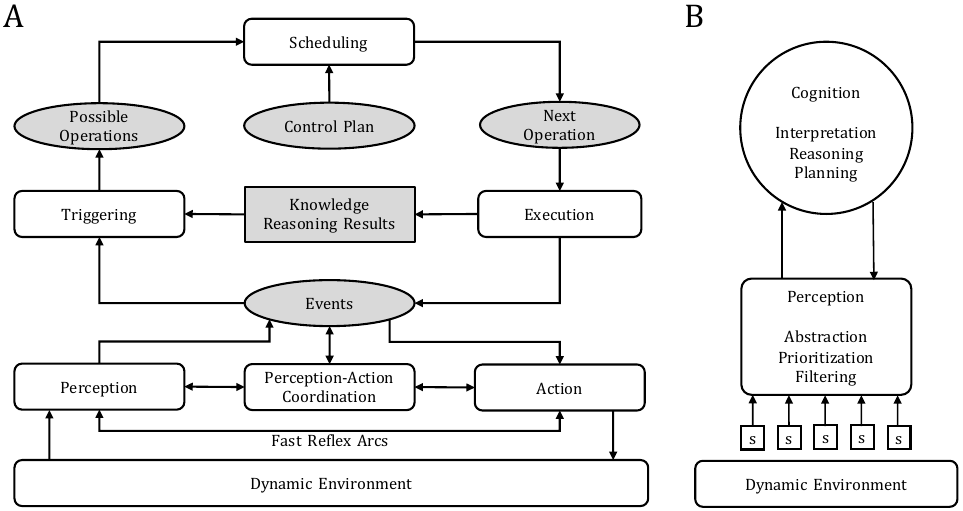}
    \caption{Active perception in adaptive intelligent systems (AIS), adapted from Hayes-Roth~\citep{hayes1995architecture}. \textbf{A:} AIS coordinates perception, reasoning, and action through adaptive control. \textbf{B:} Perception selectively acquires environmental information under cognitive guidance, forming a feedback loop between information acquisition and control.}
    \label{fig:ais}
\end{figure}

\subsection{Active perception in adaptive intelligent systems}

Task-directed control also determines what information an agent needs to acquire. A dynamic environment exposes more information than a resource-bounded agent can process continuously, so perception must select among available sources according to the agent's current activities and information needs~\citep{bajcsy1988active,bajcsy2018revisiting}. 

Hayes-Roth's architecture for adaptive intelligent systems places this selection under adaptive control~\citep{hayes1995architecture}. As illustrated in Figure~\ref{fig:ais}, cognition guides which environmental information enters subsequent reasoning and at what level of detail. Guardian provides a concrete example: as a patient's condition, incoming data, computational resources, and reasoning demands change, the system changes what information it inspects rather than applying a fixed perceptual strategy~\citep{hayes1992guardian}. Therefore, perception is not merely a fixed input stage, but part of the agent's controlled operation. 

The agent's current activities create information needs that guide perception, while the resulting observations update its assessment of the situation and thereby change subsequent reasoning and perceptual choices. We use \emph{active perception} to refer to this feedback relation between information acquisition and control. What is sensed, when it is sensed, and at what level of detail can change with the activities and conditions being served. 

Therefore, information relevance is relative to the activity that perception serves. Within a bounded task, the task objective determines the agent's current information needs. More generally, active perception couples information acquisition to the scope of activity under control: as that scope changes, what needs to be sensed can change with it.

\section{From agents in the small to agents in the large}
\label{sec:from-agent}

Persistent agency changes the unit of control. When service persists beyond an episodic task, no active task objective fully specifies what must be perceived, maintained, or adapted for future service. Therefore, a persistent architecture must organize a scope that persists across tasks, make changes to that scope observable, and convert relevant changes into bounded tasks that can affect future service. These requirements lead respectively to the notions of a \emph{long-lived setting}, active perception across task boundaries, and \emph{lifecycle tasks}.

\subsection{Episodic tasks within long-lived settings}

An episodic task ends with its bounded objective, but many of the conditions under which tasks are interpreted and performed persist beyond that objective~\citep{zhong2024memorybank,wu2024longmemeval,yu2026agentic}. Users retain preferences, artifacts and decisions persist, and rules and procedures continue to affect later tasks~\citep{tan2025prospect,rezazadeh2025collaborative,li2025memos,zhao2024expel,zhang2026memskill}. We call the persistent scope formed by the users, artifacts, constraints, decisions, and procedures whose effects survive individual tasks a \emph{long-lived setting}. Episodic tasks belong to the same setting when persistent conditions produced, assumed, or modified in one task can affect the interpretation or execution of another~\citep{maharana2024evaluating,ong2025towards,bian2026realmem}. The long-lived setting is the scope of continuing service rather than a particular memory representation; an agent's retained context represents only the parts of that setting it has captured.

A long-lived setting is distinct from a long-horizon task. A long-horizon task extends execution under one objective and ends when that objective is resolved~\citep{kwa2026measuring,zhang2026deepplanning}. A long-lived setting instead spans multiple objectives and persists beyond the completion of any one of them. Thus, \emph{long-horizon} describes the extent of execution within a task, whereas \emph{long-lived} describes the scope of coordination across tasks~\citep{zheng2025lifelongagentbench}. 

Cross-task persistence means that completing an episodic task does not resolve all conditions relevant to later tasks. Earlier decisions constrain later tasks, delayed feedback can revise prior assumptions, and artifacts, policies, tools, user needs, and reusable procedures can change over time~\citep{shen2026mem2actbench,zhang2026live,patel2026supersede,chen2026vitabench}. Therefore, previously valid context can become stale, and procedures that once succeeded can cease to fit current conditions. These changes can affect later tasks even when the current episodic objective has been successfully completed.

Cognitive language agents provide the machinery to respond once such conditions enter an active task. An \emph{agent in the large} adds a higher level of organization that treats the long-lived setting itself as the scope over which persistent conditions and cross-task relations are maintained~\citep{zhu2026your}. As in programming in the large, task-level cognition retains its own computation while a broader architecture controls properties that span individual tasks. Therefore, an agent in the large is not simply a longer-running task agent; it explicitly organizes conditions that persist beyond any individual task.

\subsection{Active perception makes the setting observable}

Within an episodic task, the objective creates the agent's information needs. It directs the agent toward relevant artifacts, memories, and environmental states, and observations are evaluated according to whether they support task completion~\citep{hayes1995architecture}. Therefore, a bounded objective provides a criterion for what information matters during task execution. 

However, across task boundaries, information can remain relevant after the objective that first exposed it has ended. A change matters at the level of the long-lived setting when it can alter persistent context, constraints, procedures, or user needs on which later tasks depend. A correction to a completed result can revise an assumption that matters to later tasks~\citep{shinn2023reflexion,patel2026supersede}; a procedure can become unreliable after its operating environment changes; and new conditions relevant to later tasks can arise before, between, or after episodic tasks~\citep{li2026proevent,yang2026contextagent}. If information enters the agent only through the observation loop of an active task, the agent cannot respond to such changes until their consequences re-enter through a later task. Therefore, task-scoped perception leaves an observability gap across task boundaries. This gap concerns what the architecture makes available for control, not how much past information it is capable of storing.

Active perception closes this gap by extending information acquisition from the current task to persistent conditions shared across tasks~\citep{hayes1995architecture,bajcsy2018revisiting}. It inspects outcomes and feedback from completed tasks, persistent context and reusable procedures, and changes in the surrounding setting~\citep{zhang2026expseek,li2026act}. Because the relevance of these sources changes as the setting evolves, sensing must adapt to what the agent currently needs to know. The agent's current understanding of the setting guides subsequent sensing, while newly perceived information updates that understanding. 

The significance of a change can also emerge only across time. A single failed action does not establish that a reusable procedure is obsolete, whereas repeated failures following an interface change provide evidence of a persistent mismatch~\citep{zhu2026your}. By relating evidence across tasks, active perception makes persistent changes observable at the level where they affect later tasks.

Perception is necessary but insufficient for persistent adaptation. The same evidence may indicate a local exception, a persistent change, or a condition outside the agent's authority~\citep{maes1994agents,horvitz1999principles,rhodes2000just}. Perception makes such conditions available for judgment, but does not determine whether or how the agent should respond. That requires a level of control that evaluates the perceived condition and turns selected changes into bounded interventions.

\subsection{Lifecycle tasks make perceived changes actionable}

A perceived condition affects later tasks only when the architecture can act on its implications. Such tasks may assess a suspected persistent change, update stale context, revise an unreliable procedure, verify that an adaptation has the intended effect, or prepare for an emerging recurring need~\citep{zhang2024agent,sun2026preference}. We call bounded tasks whose primary purpose is to maintain or improve conditions shared by future tasks a \emph{lifecycle task}. By contrast, an episodic task pursues a bounded user-facing outcome. The distinction follows from both purpose and completion criterion: an episodic task completes with respect to its bounded user-facing outcome, whereas a lifecycle task completes with respect to the evidence or change required to assess, establish, restore, or prepare a persistent condition on which later tasks depend.

Proactivity alone does not make a task lifecycle-level. An agent-initiated task remains episodic when its purpose is to produce a one-off user-facing outcome, while a user can directly request a lifecycle task when the requested change is intended to govern later tasks. For example, preparing one report is episodic, whereas correcting a preference that should govern future reports is a lifecycle task~\citep{tan2025prospect}. A single request can also contain both kinds of tasks: a request to prepare a report and use the same format for future reports induces an episodic task for the immediate report and a lifecycle task for the persistent formatting preference~\citep{wu2026ask}. Therefore, what matters is not who initiates the tasks or whether the objectives appear in the same interaction, but the purpose of each task and what its completion is intended to establish.

Lifecycle tasks reuse the same memory, reasoning, planning, and action machinery that executes episodic tasks~\citep{sumers2023cognitive,shinn2023reflexion}. Therefore, the architectural change is not a new mechanism for bounded execution, but a new scope of control over why such a task is formulated and what its completion is meant to establish. Episodic execution advances the current user-facing objective, whereas lifecycle execution produces evidence or changes intended to maintain conditions shared by later tasks.

Active perception and lifecycle tasks together add a level of organization above episodic task execution. Perception makes cross-task changes observable, lifecycle control determines which changes require intervention, and task-level cognition carries out the resulting bounded tasks. This is the architectural transition from agents in the small, whose highest-level unit of control is the episodic task, to agents in the large, whose control extends to the continuing service of a long-lived setting.

\begin{figure}[t]
\centering
\includegraphics[width=0.9\textwidth]{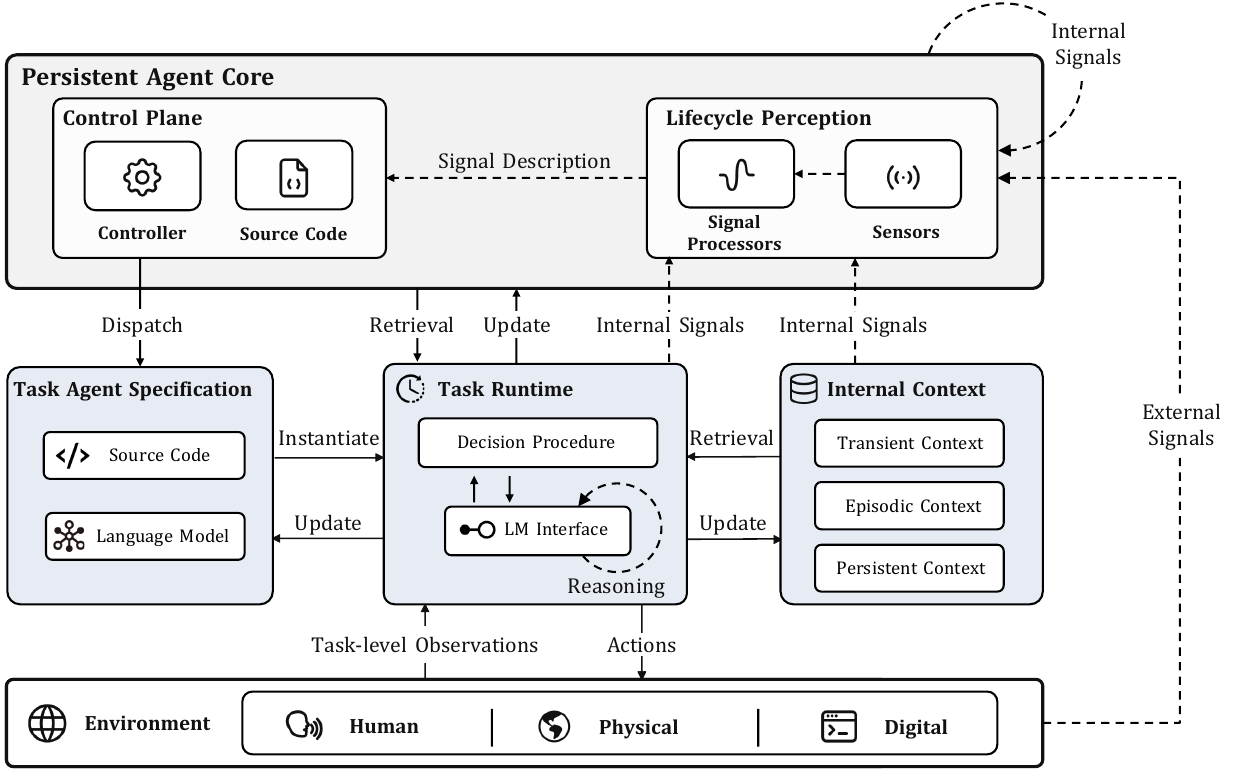}
\caption{
The conceptual perception-centered architecture for persistent agents (\textbf{Pera}). 
Pera couples active perception with task-oriented cognition, using perceived signals to formulate lifecycle tasks that update the agent and its service as the long-lived setting evolves. 
Signals can be external or internal: external signals come from human, physical, and digital environments, whereas internal signals come from a single task execution, reviews across multiple task executions and internal context memory, or perception and control themselves.
}
\label{fig:Pera-overview}
\end{figure}

\section{Pera: A \underline{Per}ception-Centered \underline{A}rchitecture for \underline{Per}sistent \underline{A}gents}
\label{sec:Pera}

We present Pera as an architecture for extending cognitive language agents~\citep{laird1987soar,sumers2023cognitive,fan2026cognitive} from episodic task execution to continuing service in long-lived settings. As shown in Figure~\ref{fig:Pera-overview}, Pera introduces a \textbf{persistent agent core} that couples lifecycle perception with a control plane. \textbf{Lifecycle perception} captures external signals from the environment and internal signals from task execution, internal context, and the agent’s own components, and interprets them to produce descriptions of service-relevant changes. The \textbf{control plane} evaluates these changes and determines whether they require further evidence or work. When adaptation is needed, it formulates a \textbf{lifecycle task package} and dispatches it to the task agent specification, which instantiates a task runtime to carry out the work through task-level decision-making.
The \textbf{task agent specification}, together with the lifecycle task package, instantiates a \textbf{task runtime} with a decision procedure specialized to the task. Through task-level decision-making, the runtime uses language-model reasoning to perform internal actions or interact with the external environment.

Pera organizes this work into two types of bounded tasks. An episodic task pursues a concrete user-facing outcome, whereas a lifecycle task establishes or restores a persistent condition on which future service depends. A task's type is determined by its purpose and completion condition rather than by whether the user or agent initiates it. Both are executed through the same task-level machinery, but lifecycle tasks additionally return evidence for lifecycle-level review, which determines whether their persistent effects should be accepted, further revised, or rolled back.

Therefore, Pera forms a recurring loop between perception, control, and task-oriented cognition: perception makes changes in the long-lived setting observable; control turns relevant changes into executable lifecycle work; task execution and review establish and assess persistent effects; and their outcomes generate further signals as the setting and agent continue to evolve. The following sections describe lifecycle perception and signals (Section~\ref{sec:Pera-signals}), lifecycle task packages (Section~\ref{sec:Pera-packages}), the shared action space (Section~\ref{sec:Pera-actions}), task-level decision-making (Section~\ref{sec:task-decision-making}), and lifecycle-level decision-making (Section~\ref{sec:lifecycle-decision-making}).

\subsection{Signals and Perception}
\label{sec:Pera-signals}

A \emph{signal} indicates an event or state change that may be relevant to future service.
The lifecycle perception captures and interprets such signals for lifecycle-level judgment, enabling proactive responses to relevant changes.

Signals and observations differ in the architectural scope of their use.
An \emph{observation} is information supplied to a task runtime during execution to inform its subsequent decisions \cite{nowe2012game,shinn2023reflexion,ding2026octobench}.
For example, a user message or feedback returned by a grounding action can serve as an observation \citep{huang2023grounded,gou2025navigating,wu2026gui}.
By contrast, a signal makes a change available to the persistent agent core as the starting point of lifecycle-level judgment.
Signals arise during a task, after a task has ended, and when no task is active.
The same event can serve both roles.
For instance, a tool error supports recovery within the current runtime while also providing evidence that a reusable procedure has become unreliable \cite{lin2026agentic,li2025review}.

According to where the indicated change occurs, signals can be divided into external and internal signals.

\paragraph{External signals.}
External signals indicate changes in the setting surrounding the agent.
Their sources parallel the physical, human, and digital environments with which cognitive language agents interact through grounding actions \citep{sumers2023cognitive}.
Physical signals can come from cameras, microphones, or sensors that measure conditions such as temperature and motion~\citep{yoon2026consensus,chai2018language,wang2025physiagent}.
Human signals reflect changes in user activity such as feedback on prior work or new needs \cite{kaufmann2023survey}.
Digital signals indicate changes in software states and artifacts, such as file-system events or updated access policies \cite{hong2024cogagent,nguyen2025gui,wang2026opencua}.

Recent proactive agents implement parts of this external perceptual interface \cite{liao2023proactive,zhang2024ask}.
ProactiveAgent monitors user activities and environmental events to predict when assistance is useful \citep{lu2025proactive}.
ContextAgent and ProAgent use open-world sensory signals, including video and audio captured by wearable devices, to infer user context and emerging needs \citep{yang2025proagent,yang2026contextagent}.
These systems show how events outside an episodic task can trigger agent assistance.

\paragraph{Internal signals.}
Internal signals indicate changes exposed by task executions or by the agent's internal context and components.
For example, execution traces expose recurring tool failures, while reviews expose source-code changes that fail to produce their intended effects~\citep{liu2026trajad,chen2026failed}.
Runtime monitoring methods already inspect task trajectories to detect risks and execution anomalies \citep{dou2026wrong,luo2025agrail,gehring2024rlef,dou2024stepcoder}.

Changes in internal context can likewise generate signals.
The lifecycle perception and scheduled tasks inspect how retained context evolves and identify changes relevant to continuing service.
For example, accumulated information that reveals an emerging user interest triggers proactive assistance, whereas recurring stale entries expose deficiencies in context maintenance and can lead to a lifecycle task that revises the relevant procedure or source code~\citep{chao2026stale,xu2026mem,zhang2025survey,yang2026beyond}.

Signals represent either individual events or conditions inferred from related events \cite{cugola2012processing}.
For example, the creation of a directory can produce a single-event signal.
A single failed tool call does not establish a persistent problem \cite{chen2025learning}.
Repeated failures following an interface change instead provide signals that a reusable procedure has become obsolete.
Signal aggregation allows the lifecycle perception to recognize conditions that are difficult to identify from one event alone \citep{wang2025probguard,chowa2026language}.

\begin{figure}[t]
\centering
\includegraphics[width=0.9\textwidth]{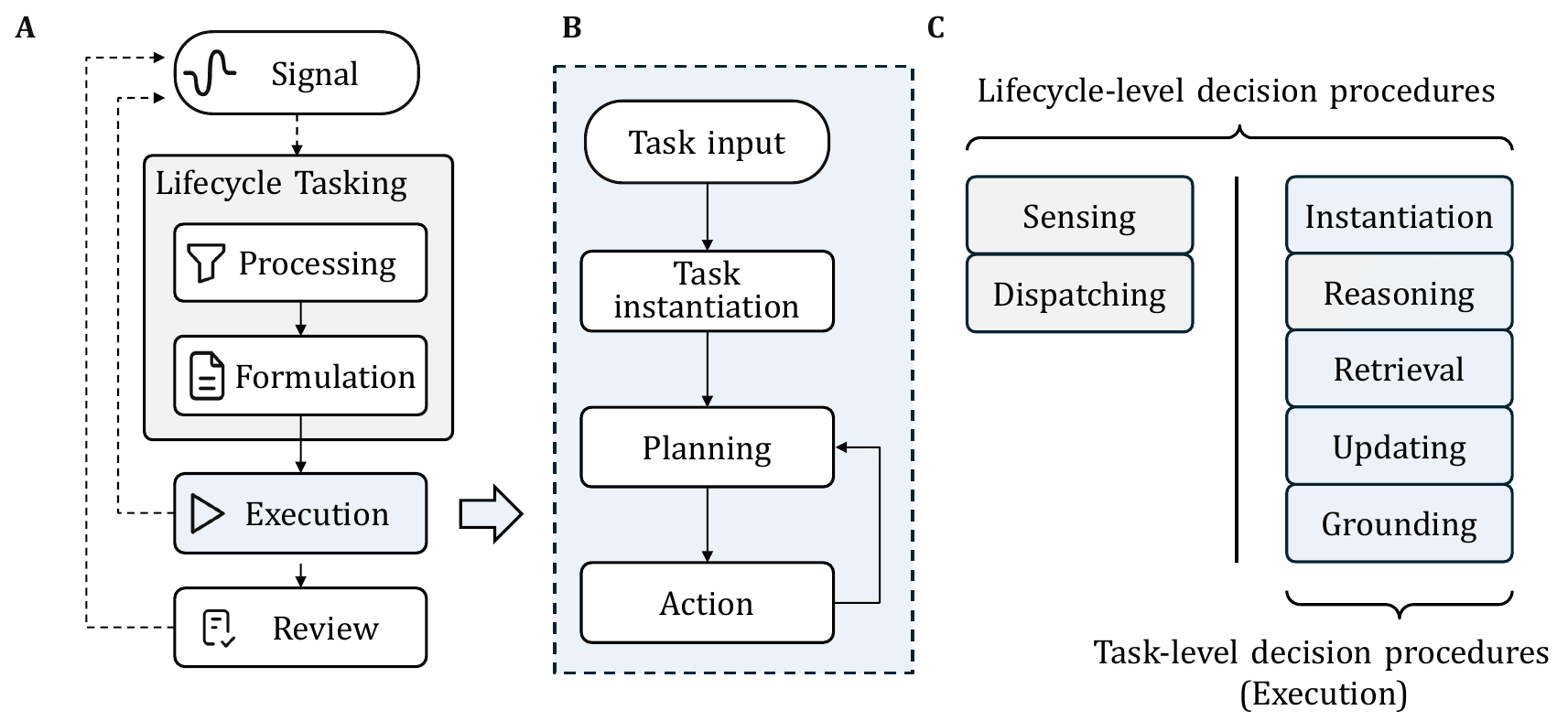}
\caption{
Lifecycle-level and task-level decision procedures and the action space in Pera. 
\textbf{A:} Lifecycle-level decision-making actively senses and interprets signals, determines whether to formulate lifecycle tasks, executes them through task-level decision-making, and reviews their execution and outcomes. 
Processing and formulation can invoke task-level execution when their work requires it, with the resulting outputs or signals feeding back into the ongoing lifecycle-level decision-making process.
Execution and review can in turn generate new signals. 
\textbf{B:} Task-level execution begins with a task input, either a task prompt or a lifecycle task package. The agent instantiates a task runtime using the task agent specification and, when available, the task-specific procedure in the package, and then runs a repeated planning--action cycle until completion or failure. 
\textbf{C:} Sensing and dispatching are specific to lifecycle-level decision procedures, while reasoning can support both lifecycle-level signal processing and formulation and task-level planning. During task-level decision-making, planning selects internal actions, including retrieval and updating, or external grounding actions. Their outcomes can in turn become signals captured through sensing.
}
\label{fig:Pera-procedure_action_space}
\end{figure}

\paragraph{Lifecycle perception.}
The lifecycle perception comprises two components for capturing and interpreting signals.
Specifically, sensors detect signals, while signal processors interpret them to produce descriptions of what has happened and make these descriptions available to the control plane.

Sensors can take physical or software forms.
Physical sensors capture measurements or multimodal inputs from the surrounding environment \citep{driess2023palm,reed2022generalist,mu2023embodiedgpt}.
Software sensors can receive events from operating systems and applications or monitor the agent's internal state and activity \citep{agashe2025agent,wu2024copilot,xie2024osworld}.
Sensors capture signals either as events occur or through periodic inspection of relevant state \citep{ziafati2013event,rivkin2023sage}.
The resulting inputs can describe an individual event or summarize changes observed over an interval.

Signal processors interpret the captured signals and describe the indicated condition with sufficient specificity for subsequent judgment.
Signal processing suppresses irrelevant inputs and combines related signals when a condition becomes evident only over time.
A sequence of temperature signals, for example, can be interpreted as indicating that the workspace temperature has remained above a specified threshold \citep{rivkin2024aiot,rivkin2023sage,gyllstrom2006sase}.
Signals collected from several task executions can similarly be summarized as a recurring interface failure.
For changes that require multimodal understanding, retrieval, or tool use, signal interpretation can itself be packaged as a lifecycle task and executed by a task runtime~\citep{majumdar2024openeqa}.

The lifecycle perception passes the resulting description to the control plane, which evaluates the described change against the current states of the agent and its environment and determines whether further evidence or a lifecycle task is required.
If the description is incomplete, further sensing or verification can itself be packaged as a task.
Section~\ref{sec:lifecycle-decision-making} describes this subsequent process.

Although existing work has explored parts of this perceptual process, external and internal sources can produce many potentially irrelevant signals, making appropriate choices about sensing intensity and granularity critical \cite{zaky2020active}.
Accurately interpreting signals and identifying their underlying causes remain a core challenge for lifecycle perception.
We revisit these challenges and their design implications in Section~\ref{sec:actionable-insights}.

\subsection{Lifecycle Task Packages}
\label{sec:Pera-packages}

By analyzing an interpreted signal that describes what has happened, the control plane determines what work should follow and constructs a \emph{lifecycle task package}.
The resulting package converts the required work into a form that can be dispatched and executed.
As the interface between the control plane and task-level execution, it specifies the task objective and procedure, together with operational requirements such as context acquisition, failure recovery, and review \cite{hong2024metagpt,gao2024agentscope}.

At the center of the package is a clear task objective describing what the runtime should accomplish and what persistent effect should remain after execution.
The package also defines a completion criterion that specifies the evidence the runtime must produce before it stops and returns its result.
For example, suppose a computer-use agent repeatedly fails to submit expense reports after the reporting website changes its interface \cite{yan2025mcpworld,ishmam2026timewarp}.
In this example, the package instructs the runtime to inspect the updated interface and revise the reusable submission procedure.
Its completion criterion can require the runtime to return the revised procedure together with a trace showing successful submission of a test expense report through the updated interface.

To make the objective executable, the package can also provide or reference a procedure that guides execution.
The procedure can reuse an existing workflow, introduce task-specific code, or equip the runtime with a new tool for grounding in the environment \citep{wang2023voyager,nguyen2024dynasaur,cai2024large}.
The package also supplies the context needed for execution or specifies how it should be retrieved.
In the example above, it can provide the failed task traces and the current submission procedure, then instruct the runtime to inspect the changed interface before proposing a revision.

Moreover, lifecycle tasks that modify reusable procedures or source code can affect later services when the update fails \cite{moll2026grasp,ren2026self}.
Therefore, for tasks that modify persistent components, the package includes recovery mechanisms that preserve a stable version and restore it when the proposed change fails~\citep{li2025generator,hosek2013safe,giuffrida2013back}.
In the running example, the runtime retains the previous submission procedure and restores it when the revision fails validation.
The package can also restrict modification to specified components or require additional authorization before consequential actions are executed \citep{wang2025agentspec,zhang2026you,south2025authenticated,zhu2025miniscope}.

The runtime determines task completion, whereas the lifecycle level determines whether the returned change becomes part of future service. Therefore, the package further specifies a review criterion stating what must hold before a proposed change is accepted~\citep{zhang2026self,moll2026grasp}.
Review extends beyond completion: it examines evidence that the runtime was not required to produce, in particular whether the change preserves behavior the agent already supported.
In the running example, completion requires one successful submission through the updated interface, whereas review additionally replays the revised procedure on the report types that succeeded before the change, and only then does the revision replace the stored procedure \cite{lou2024automated}.
The assessment work can be performed within the original execution procedure, or dispatched as a separate task when an independent evaluator is needed~\citep{singhi2026think,zhuge2024agent,chan2024chateval}.
The acceptance decision itself remains with the control plane.

In practice, these elements do not prescribe a fixed package structure.
The control plane adapts the package's contents and level of detail to the difficulty and expected impact of each lifecycle task.
If lifecycle tasks repeatedly expose deficiencies in package construction, revising the package-formulation procedure can itself become a lifecycle task \cite{zhang2025aflow,hu2025automated}.

\subsection{Action Space}
\label{sec:Pera-actions}

Pera defines an action space covering operations within the agent and between the agent and the external environment.
It comprises \emph{sensing}, \emph{dispatching}, \emph{instantiation}, \emph{reasoning}, \emph{retrieval}, \emph{updating}, and \emph{grounding}, as shown in Figure~\ref{fig:Pera-procedure_action_space}.
Reasoning, retrieval, updating, and grounding build on cognitive language agents' internal and external actions \citep{sumers2023cognitive,wang2024survey}.
Pera further incorporates sensing, dispatching, and instantiation for ongoing perception and lifecycle-level coordination.

\paragraph{Sensing.}
Sensing proactively monitors external and internal sources for signals relevant to continuing service.
Depending on signal source and monitoring objective, sensing can respond to individual events or periodically inspect a relevant state.
For external signals, ProactiveAgent~\citep{lu2025proactive} monitors user activities and environmental events; ContextAgent~\citep{yang2026contextagent} and ProactiveVA~\citep{zhao2025proactiveva} infer user context and emerging needs from multimodal or interface observations.
For internal signals, runtime monitors inspect task trajectories to detect unsafe states, anomalous steps, and policy violations~\citep{wang2025probguard,liu2026trajad}.

Sensing mechanisms also differ in how much computation they use and how signals are collected over time.
Sensing can adaptively allocate perception, using lightweight observations for routine monitoring while invoking costlier perception when additional evidence is needed~\citep{ren2024explore,zaky2020active}.
Moreover, long-running sensing requires sustained observation, as relevant conditions can emerge only across events or state changes at different moments~\citep{maldaner2026sentinelbench,ziafati2013event}.

While recent work has explored different signal sources and sensing mechanisms, how sensing should adapt as the environment and agent state change remains understudied in recent language agents.
Persistent agents need to expand or revise the sources they monitor, adjust sensing frequency and granularity, and determine when related events should be considered together~\citep{ren2024explore}.

\paragraph{Dispatching.}
Dispatching sends a formulated lifecycle task package from the control plane to the agent specification and determines when the package enters execution.
The control plane chooses among immediate dispatch, delayed execution, and waiting for resources, dependencies, or authorization \cite{yu2025dyntaskmas,guo2026saga}.
At the system level, AIOS~\citep{mei2024aios} schedules concurrent agent requests and manages their access to shared model and tool resources.
For dependent-subtask workflows, DynTaskMAS~\citep{yu2025dyntaskmas} uses dynamic task graphs to support asynchronous and parallel execution.

Dispatching can account for resource availability, task dependencies, task priority, and execution state \cite{xie2025ai,zheng2026agentcgroup}.
Agent.xpu~\citep{wei2025agent} maintains separate reactive and proactive workload priority queues, and Agent libOS~\citep{zhang2026agent} supports scheduling, authorization, and resumption of long-running agents.
For persistent agents, dispatching must also consider how lifecycle tasks that maintain or improve future service should be scheduled alongside episodic tasks with immediate response requirements.

\paragraph{Instantiation.}
Instantiation turns the reusable agent specification into an executable task runtime for a particular task.
The action binds the task input to the general decision procedure.
The agent specification encodes this procedure in the agent's source code and language-model parameters~\citep{sumers2023cognitive,gao2024agentscope,wu2023autogen}.
A common implementation makes this transition implicit: a task instruction directly initiates the agent's decision procedure~\citep{yao2022react,huang2022language,singh2022progprompt}.
Pera explicitly models this transition as an instantiation action, analogous to creating a running instance from a reusable program.
For a lifecycle task, instantiation applies the package's task-specific procedure to supplement or constrain the general decision procedure before execution.

\paragraph{Reasoning.}
Reasoning allows a task runtime to process the information currently available to generate new information.
Unlike retrieval, which brings stored information into the runtime, reasoning transforms the information already available \citep{sumers2023cognitive,wei2022chain,yao2023tree,hao2023reasoning}.
It can interpret the most recent observation and revise its plan \citep{wang2023voyager,skreta2024replan}, distill insights from execution trajectories \citep{zhao2024expel,ouyang2026reasoningbank}, or synthesize information retrieved from internal context \citep{park2023generative,lee2024human}.
It can also identify additional context to retrieve, evaluate which grounding action to take, and guide updates to internal context or agent components \cite{hao2023reasoning}.
In Pera, reasoning can also support sensing and signal processing by helping the signal processor interpret captured signals or analyze the resulting signal description to clarify what it indicates and inform the formulation of a subsequent lifecycle task package \cite{zeng2022socratic}.
Active-perception agents similarly use reasoning or reflection to determine when and where to acquire observations \citep{zhang2024proagent,lee2025sensible,wang2026active}.
When reasoning reveals that necessary evidence is unavailable because the current sensing procedure is insufficient, the task execution can produce an internal signal that prompts the control plane to formulate a lifecycle task for revising the relevant sensing or signal-processing procedure.

\paragraph{Retrieval.}
Retrieval reads relevant information from internal context for decision-making \cite{park2023generative,packer2023memgpt,wang2024agent}.
Pera organizes internal context into transient, episodic, and persistent context.
Transient context is task-local information temporarily stored outside the language model's context window and retrieved as needed during the same execution.
Episodic context retains knowledge and experience accumulated through task executions for later use, while persistent context retains information relevant to continuing service in the long-lived setting.
Depending on context representation, retrieval uses rule-based filtering, sparse lexical matching such as BM25 \citep{robertson2009probabilistic}, dense embedding retrieval \citep{karpukhin2020dense}, or neural ranking and late interaction \citep{nogueira2019passage,khattab2020colbert}.

Recent agent-memory systems retrieve different forms of accumulated context~\citep{hu2025memory}, including salient information extracted from multi-session interactions in Mem0 \citep{chhikara2025mem0}, records organized as dynamically linked notes in A-MEM \citep{xu2026mem}, and textual strategies distilled from successful and failed experiences in ReasoningBank \citep{ouyang2026reasoningbank}.
In Pera, retrieval is limited to internal context, since the general decision procedure is applied through instantiation, while information from the external environment is acquired through grounding.

\paragraph{Updating.}
Updating changes the agent's internal state or implementation.
Language agents can update internal context by retaining task experience, consolidating knowledge, and revising accumulated information \citep{zhong2024memorybank,li2025memos,ouyang2026reasoningbank}.
At the agent-specification level, model parameters can be updated through diverse learning methods~\citep{zeng2024agenttuning,sun2025seagent,hu2026seal}; source-code elements and reusable procedures can be added, revised, or removed to improve task-level decision-making \citep{hu2025automated,shang2025agentsquare,ni2026trace2skill}.

Viewed through Pera, recent self-evolving agents mainly update the source code, tools, and reusable procedures in the agent specification~\citep{lee2026meta,lin2026agentic,zhang2026darwin}.
For example, STOP~\citep{zelikman2023self} improves an LM-based decision procedure encoded in source code; SICA~\citep{robeyns2025self} and other recent work~\citep{yin2025godel,zhang2026darwin} modify agent source code; and some agent frameworks~\citep{zhang2026self,liu2026adaptive} extend updating to broader execution-harness components.

In Pera, updating can also modify the persistent agent core.
For example, a deficiency in sensors or the controller can motivate a lifecycle task that revises the affected component itself.
Updating can directly revise software sensors and signal processors \citep{murphy1999handling,peters2024robot}.
Changes to physical sensors are carried out through grounding: a task runtime can use embodied action to recalibrate, repair, or replace a sensor, or proactively ask the user to make such a change.
Such changes can alter the physical sources from which the lifecycle perception captures future signals \cite{murphy1999handling,peters2024robot}.
Updating can then revise the corresponding software interface or grounding procedure to incorporate a changed or newly introduced device.
Like updates to model parameters and source code, updates to the persistent agent core are risky because errors can affect perception~\citep{shao2026your}.

\paragraph{Grounding.}
Grounding allows a task runtime to interact with the external environment through available mechanisms, such as software tools, communication channels, and embodied action \citep{zhou2024webarena,qin2024toolllm,shao2026collaborative}.
Feedback from this interaction is returned to the runtime as an observation for subsequent decisions.
Unlike sensing, which captures service-relevant changes, grounding is performed by a task runtime to advance its current task.

Recent work has explored how environmental interaction is incorporated into agent decision-making \citep{yao2022webshop,zhang2025appagent}.
Existing language-agent frameworks also consider digital environments, interactions with humans, and physical environments as broad forms of external grounding \citep{sumers2023cognitive,zhou2024webarena}.

In Pera, grounding interactions across these environments can reveal information about the user or long-lived setting that matters beyond the current task.
For example, user edits can reveal latent preference patterns, while direct feedback can indicate a new requirement or a setting change~\citep{sun2026preference,gao2024aligning,shao2026collaborative}.
The lifecycle perception can also capture information produced or revealed during grounding as an external signal, allowing the control plane to determine whether to formulate a lifecycle task that updates service procedures.
Recent agents have begun learning latent preferences from user edits and behavioral histories \citep{gao2024aligning,wang2026me} and adapting to user desires across embodied or repeated interactions \citep{wang2025strangers,mehri2026multisessioncollab}.

Sections~\ref{sec:task-decision-making} and~\ref{sec:lifecycle-decision-making} describe how these actions are organized within task-level and lifecycle-level decision-making, respectively.

\subsection{Task-Level Decision-Making}
\label{sec:task-decision-making}

Task-level decision-making is the process through which a task runtime carries out an individual task.
A task enters task-level execution either directly through a task request or through a dispatched lifecycle task package.
As illustrated in Figure~\ref{fig:Pera-procedure_action_space}, it first enters instantiation, after which the resulting runtime proceeds through a repeated cycle of planning and action \cite{de2020bdi,sumers2023cognitive,pmlr-v205-huang23c,yao2022react,Georgeff1989Decision}.
For work dispatched through the control plane, the task-level decision-making process realizes the execution stage.

\paragraph{Task instantiation.}
Task instantiation applies the task input to the agent specification to create a task runtime \cite{shoham1993agent,laird1987soar}.
For a directly provided task request, it initializes the general decision procedure defined by the agent specification with that request \cite{sumers2023cognitive}.
When the input is a lifecycle task package, its task-specific procedure and operational requirements further supplement or constrain the general procedure.
Notably, language-model computation occurs during execution of this instantiated procedure, while the LM interface is depicted separately in Figure~\ref{fig:Pera-overview} to make explicit how the procedure exchanges prompts and responses with the language model.

\paragraph{Planning.}
Planning produces and selects the next action given the task objective and the runtime's current task state \cite{yao2023tree,hao2023reasoning,zhou2023language}.
Within planning, the instantiated decision procedure uses reasoning actions to generate candidate actions and assess their suitability or anticipated consequences before selecting what to execute next \cite{yao2022react,pmlr-v205-huang23c}.
If additional information is needed, planning selects retrieval or grounding as the next action.
The result of that action can then inform a later planning cycle.

Recent agent frameworks differ in how candidate actions are generated, whether and how they are evaluated, and how far the procedure looks ahead before selecting one \cite{yao2023tree,wang2023describe}.
For example, Soar's~\citep{laird1987soar} decision cycle proposes and evaluates operators before one is selected for application, while CoALA~\citep{sumers2023cognitive} describes language-agent planning in terms of proposing, evaluating, and selecting candidate actions.
ReAct~\citep{yao2022react} produces the next action incrementally from the evolving interaction history, whereas SayCan~\citep{ahn2022can} selects among predefined robotic skills by combining language-model predictions with estimated feasibility.

More recent work makes the comparison of alternatives before execution increasingly explicit. 
Tree of Thoughts~\citep{yao2023tree} generates and evaluates multiple candidate reasoning states, while RAP~\citep{hao2023reasoning} and LATS~\citep{zhou2023language} use search and value estimates to compare longer reasoning or action trajectories before selection.
A complementary line of work~\citep{Song2022LLMPlanner,dagan2023dynamic,munoz2025chathtn} imposes explicit structure on planning, for example by revising plans using execution feedback, integrating language models with symbolic planners, or organizing actions into hierarchical or behavior-tree representations.

\paragraph{Action.}
The action stage executes the retrieval, grounding, or updating operation selected during planning \cite{schick2023toolformer,Wang2024ExecutableCA}.
The outcome is returned to the runtime as an action result and contributes to its current task state for the next planning cycle.

Language agents represent and execute selected actions in different ways \cite{patil2024gorilla}.
For example, Toolformer~\citep{schick2023toolformer} and Gorilla~\citep{patil2024gorilla} encode external API calls in model outputs, whereas CodeAct~\citep{Wang2024ExecutableCA} represents actions as executable code interpreted by a code runtime.
Some agent frameworks~\citep{qin2025ui,feng2026retool} interleave language-model reasoning with tool execution and incorporate returned results into subsequent decisions, while others translate model decisions into direct interface operations such as keyboard and mouse actions.
Planned steps can also be delegated to specialized execution components \citep{Lu2023ChameleonPC,patil2024gorilla,agashe2025agent}. Execution outcomes can then be compared with their expected effects to detect failures and guide subsequent recovery \citep{zhang2026don}.

Planning and action repeat until the task reaches its completion condition or the runtime can no longer continue under the current procedure and constraints.
A failed action result can trigger a retry or a revised course of action in a later cycle.
The runtime returns task results and artifacts on completion, or available evidence of failure when it cannot continue.
When these outputs or execution traces reveal a service-relevant condition, the lifecycle perception can capture the evidence as an internal signal.

\subsection{Lifecycle-Level Decision-Making}
\label{sec:lifecycle-decision-making}

Lifecycle-level decision-making is the process through which Pera interprets external and internal signals and organizes the work needed to sustain and improve service in a long-lived setting \cite{Kephart2003the,garlan2004rainbow,park2023generative}.
As illustrated in Figure~\ref{fig:Pera-procedure_action_space}, it proceeds through four recurring stages, i.e., \textit{processing}, \textit{formulation}, \textit{execution}, and \textit{review}.
\textit{Processing} transforms an incoming signal into a description of the perceived condition, expressed in a domain-specific language that the control component can reason over.
Based on this interpretation, \textit{formulation} defines lifecycle tasks and constructs the corresponding task packages described in Section~\ref{sec:Pera-packages}, supplying the content needed for their \textit{execution} and \textit{review}.

When existing procedures are insufficient, formulation generates task-specific procedure code and includes it in the package.
Each package is dispatched to a task agent specification and applied during instantiation, where the task-specific procedure in the package supplements or constrains the general decision procedure defined by the specification.
This creates a task runtime with a decision procedure specialized to the lifecycle task. The runtime executes the task through task-level decision-making and returns results for review against the applicable requirements.
Execution and review can produce further signals, allowing the process to continue as the setting or the agent changes.

\paragraph{Processing.}
Signal processors aggregate the signals acquired by the sensors described in Section~\ref{sec:Pera-signals} and transform them into descriptions of perceived conditions, expressed in a domain-specific language.
Different systems instantiate this interface according to their sensing domains.
For example, recent wearable and proactive-agent systems~\citep{yu2025sensorchat,yang2026contextagent,zhang2026sensorlm} translate continuous multimodal sensor streams into textual descriptions or align the streams with textual descriptions and contexts for language-based reasoning.
Embodied systems~\citep{driess2023palm,gu2024conceptgraphs,gao2026arcadia} similarly incorporate visual, proprioceptive, thermal, and other sensor modalities into language models or summarize multisensory observations as semantic scene representations and natural-language explanations.
The resulting description is passed to formulation.

\paragraph{Formulation.}
Formulation starts from the interpretation produced by processing and determines whether lifecycle work should follow and, if so, how it should be organized.
The control plane uses a language model as its general reasoning mechanism and plays a role analogous to an operating-system core, handling basic coordination directly while delegating extended work to task runtimes through lifecycle tasks.
For each task, it constructs the corresponding package with the elements described in Section~\ref{sec:Pera-packages}.
Among these elements, the task-specific procedure is the one produced during formulation itself, and it supplements or constrains the general decision procedure defined by the task agent specification during instantiation.

The task-specific procedure can be coded directly by the control plane, as illustrated by agents that dynamically construct executable actions or tools when predefined capabilities are insufficient \citep{Suris2023ViperGPTVI,liang2023code,qian2023creator}.
Alternatively, its construction or validation can be formulated as another lifecycle task when direct coding exceeds the control plane's computational capacity or requires extended work \cite{hong2024metagpt,hu2025automated}.
Moreover, when the available signals are insufficient or their interpretation is inadequate, the control plane can also formulate lifecycle tasks to extend or revise the relevant sensing and processing components, such as by adapting sensor configurations or dynamically orchestrating modality-specific perception tools \citep{tao2025active,chen2026task,Wu2023VisualCT}.
Once formulation is complete, the control plane dispatches each package to a task agent specification for instantiation and execution.

\paragraph{Execution.}
Both episodic and lifecycle tasks are carried out through the task-level decision-making process described in Section~\ref{sec:task-decision-making}.
To maintain Pera's architectural abstraction, a lifecycle task originating from a user request is also formulated by the control plane as a lifecycle task package and is then dispatched for execution.
The execution process can produce new internal signals, such as execution errors.
These signals provide evidence for formulating lifecycle tasks that revise deficient components, such as revising source code based on failures \citep{zhang2026self,shinn2023reflexion,Zhuge2024LanguageAA}.
Additionally, execution results can produce internal signals, such as indications that an outcome fails to meet expectations \cite{Yang2023InterCode,Guo2024UsingGA,gou2024critic}.
These signals are examined and consolidated in the subsequent review stage and are also captured by the lifecycle perception.

\paragraph{Review.}
In the review phase, the agent examines execution-related evidence, such as the execution trajectory, task outcome, and resulting state, to surface internal signals \cite{madaan2023self,zhuge2024agent,shinn2023reflexion}.
Review remains a distinct lifecycle-level stage, although parts of the review process are carried out through additional task-level execution.

For an episodic task, review evaluates the immediate execution, such as whether the task objective was achieved, whether the result is correct, and whether the resulting state is consistent with the user's request~\citep{huang2026beyond,zhang2026don}.
For a lifecycle task, the task-specific procedure in the package can further specialize this review.
Beyond what is examined for an episodic task, review of a lifecycle task also determines whether the intended persistent condition was established and whether any resulting update should be accepted, further revised, or rolled back \cite{6035728,6606623}.
Review can also extend over time or across tasks, such as by scheduling a monitoring task to inspect logs after a modified component has been in use for some time, or by identifying user behaviors from repeated requests across task trajectories \citep{31339563134015,31043223104329,pan2026retrospective}.

Review uses methods such as reflection, active probing of the environment or test generation to verify a task outcome, or assessment of a harness revision against execution evidence accumulated over subsequent tasks \citep{madaan2023self,gou2024critic,10650101065036}.
When the available evidence is insufficient, review can be grounded in additional user input, such as by summarizing the unresolved issue and proactively asking a clarification question \citep{zhang2024ask,wang2025learning,Tellex2014AskingFH}.

The lifecycle perception can capture the findings produced by review. These findings can ground further interaction with users, motivate lifecycle tasks, update the agent's understanding of the user, or lead to revisions of its own components.
Review and the lifecycle perception thus complete the feedback loop from task-level execution back into lifecycle-level decision-making, allowing evidence from individual tasks and across tasks to shape future service.

\begin{figure}
    \centering
    \includegraphics[width=0.9\linewidth]{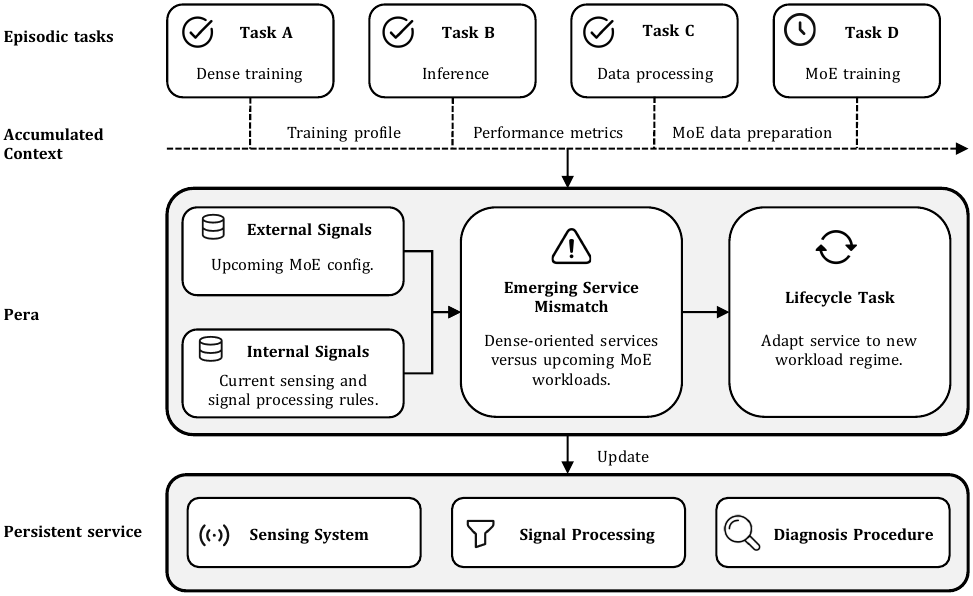}
    \caption{
Illustration for a long-lived model-development setting.
Interdependent episodic tasks accumulate context and introduce new workload requirements.
Pera detects a mismatch between the evolving setting and the current reliability service, dispatches a lifecycle task to revise shared sensing and diagnostic procedures, and preserves the update for future episodic tasks.
}
    \label{fig:case_study}
\end{figure}

\section{Case Study: Persistent Reliability for an Evolving Model-Development Workload}
\label{sec:case_study}

We use a model-development project on a shared computing cluster to illustrate how context, perception, and control shape continuing agent service. The same user performs a sequence of related but independently bounded workloads, while project artifacts, the computing environment, and the agent's service procedures persist across them. This setting separates the completion of individual workloads from the maintenance of the shared conditions that support them over time.

Figure~\ref{fig:case_study} summarizes the case. The upper sequence shows four episodic tasks whose artifacts, measurements, and operating experience accumulate as persistent context. Pera relates this context to external and internal signals and evaluates whether the existing reliability service remains appropriate as the workload changes. The transition toward Mixture-of-Experts (MoE) training exposes a mismatch between the upcoming workload and the existing sensing and diagnostic procedures, which triggers a lifecycle task that updates these shared procedures for subsequent workloads.

\subsection{A Long-Lived Model-Development Setting}
\label{sec:model_development_setting}

Consider a user developing a model through four successive stages on the same computing cluster. Dense training first produces a base model and its checkpoints. Inference and evaluation then produce performance measurements and error cases. These results guide a data-processing task that filters, transforms, and reorganizes training data. The processed data and earlier model artifacts are finally used in MoE training.

Each stage is an episodic task with its own objective and completion condition, but the stages belong to the same long-lived setting because their artifacts and operating conditions persist across task boundaries. Dense training produces the model consumed by inference; inference produces evidence that guides data processing; and data processing produces the dataset consumed by later training. Resource configurations, performance profiles, diagnostic results, user preferences, and reusable operating procedures also persist across the sequence.

By the end of data processing, the setting already shows signs of the next change. Accumulated project context records the characteristics of dense training, inference, and data processing, while the next model configuration and resource plan indicate a transition toward MoE workloads. At the same time, the agent's monitoring baselines and diagnostic procedures still encode assumptions from the earlier workload regime. Figure~\ref{fig:case_study} represents this condition as the emerging service mismatch: the workload is changing faster than the reliability service that interprets it. No current episodic task has failed; the relevant condition is that the persistent reliability service is no longer aligned with the workloads it must support.

We use this mismatch to compare the architectures below. The comparison asks whether each architecture preserves information across episodic tasks, perceives changes whose significance extends beyond the current objective, and turns those changes into work that maintains the service procedures shared by future tasks.

\subsection{Architectural Comparison}

Table~\ref{tab:agent-architecture-comparison} maps four representative systems and Pera onto the perception, control, and context dimensions introduced in Sections~\ref{sec:from-agent} and~\ref{sec:Pera}. Each architecture is considered under the same evolving project. This comparison isolates what each architecture preserves across the sequence, what changes it perceives, and what kind of work those changes produce.

\begin{table}[t]
\centering
\small
\setlength{\tabcolsep}{4.2pt}
\renewcommand{\arraystretch}{1.18}
\caption{
    Representative agent architectures organized by perception, control,
    and context. The comparison considers how each architecture supports a
    sequence of related episodic tasks as the computational requirements of
    the long-lived setting evolve.
}
\resizebox{\linewidth}{!}{
\begin{tabular}{lccccc}
\toprule

    \multirow{2}{*}{\textbf{System}}
    & \multicolumn{2}{c}{\textbf{Perception}}
    & \multicolumn{2}{c}{\textbf{Control}}
    & \multirow{2}{*}{\textbf{Context}} \\

    \cmidrule(lr){2-3}
    \cmidrule(lr){4-5}

    & \textbf{Scope}
    & \textbf{Source}
    & \textbf{Tasking}
    & \textbf{Decision-making} \\
    \midrule

    Plan-and-Act \cite{erdogan2025plan}
    & observation
    & external
    & episodic
    & task-level
    & transient \\

    A-MEM \cite{xu2026mem}
    & observation
    & external
    & episodic
    & task-level
    & episodic \\

    PaLM-E \cite{driess2023palm}
    & observation
    & external
    & episodic
    & task-level
    & transient \\

    ContextAgent \cite{yang2026contextagent}
    & observation/signal
    & external
    & proactive episodic
    & task-level
    & transient/episodic/persistent \\

    \textbf{Pera}
    & \textbf{observation/signal}
    & \textbf{external/internal}
    & \textbf{episodic/lifecycle}
    & \textbf{task-level/lifecycle-level}
    & \textbf{transient/episodic/persistent} \\

    \bottomrule
\end{tabular}}

\label{tab:agent-architecture-comparison}

\end{table}

\paragraph{Task-centered agent: Plan-and-Act.}
Plan-and-Act \citep{erdogan2025plan} organizes external observations and transient context around the current episodic objective. It plans and executes each stage when that stage is presented as a task: configure dense training, inspect an inference run, carry out data processing, or execute an MoE workload.

Its control ends at the boundary of the active task. Information retained during one execution serves that execution, and the next workload begins under a new task-level objective. Therefore, project-level dependencies and changes in reliability requirements remain outside the architecture's persistent control and must be rediscovered within later episodic tasks.

\paragraph{Memory-augmented agent: A-MEM.}
A-MEM \citep{xu2026mem} adds episodic context, allowing experience from one stage to remain available in later stages. Dense-training behavior, inference results, diagnostic observations, and prior user feedback remain retrievable when the project moves to data processing or later training. This creates continuity across the episodic sequence.

However, the retained experience remains an input to task-level decision-making. The emerging MoE workload becomes operationally relevant when a later task retrieves and uses that context. Memory preserves how the project reached its current state, but does not turn the changing workload regime into work that maintains the reliability service itself.

\paragraph{Sensor-grounded agent: PaLM-E.}
PaLM-E \citep{driess2023palm} extends the perception available within an episodic task through external sensor observations. In the model-development setting, this grounding provides information about accelerator, CPU, storage, and other system conditions relevant to the workload currently being executed.

These observations remain organized around the active task. They improve the agent's understanding of the current workload, but do not preserve the accumulated project history or track how reliability requirements change across workload stages. Therefore, PaLM-E extends what the current task observes, whereas A-MEM extends what later tasks remember.

\paragraph{Context-aware proactive agent: ContextAgent.}
ContextAgent \citep{yang2026contextagent} combines persistent context with continuing external perception. Therefore, the transition toward MoE workloads becomes visible before the user starts the next training task. The agent relates the upcoming configuration and resource plan to accumulated project context and proactively initiates an episodic service, such as assessing the expected workload or informing the user that operating conditions have changed.

Its control scope remains episodic. The perceived change produces another service task directed at the emerging workload, while the sensing and diagnostic machinery shared across workload stages remains outside the maintained service state. Proactivity changes when assistance begins, but does not extend control from individual service tasks to the persistent procedures that support them.

\paragraph{Persistent agent: Pera.}
Pera extends the scope of control from individual workloads to the reliability service shared across them. In Figure~\ref{fig:case_study}, external signals describe the transition toward MoE training, including the upcoming model configuration and resource plan, while internal signals expose the assumptions encoded in the agent's existing sensing and diagnostic procedures. Combined with the accumulated context from earlier workloads, these signals reveal an emerging service mismatch: the project is moving toward a new workload regime, but the reliability service remains calibrated to the previous one.

This mismatch is significant because it concerns future service rather than the success of the current episodic task. Therefore, Pera turns it into a lifecycle task that revises the shared sensing, signal-processing, and diagnostic procedures. As illustrated in Figure~\ref{fig:case_study}, the lifecycle task does not replace or extend the upcoming MoE training task. Instead, it updates the persistent service used across workloads so that normal and abnormal behavior can be interpreted according to the new workload regime while preserving support for earlier stages.

Once validated, the revised procedures become part of the agent's continuing service state. The subsequent MoE training remains an ordinary episodic task, but it is now served by a reliability service that has already adapted to the evolving project. Therefore, Pera connects two capabilities that the preceding architectures separate: signals make cross-task changes observable, and lifecycle tasks turn those changes into persistent adaptations that benefit future episodic tasks.

\section{Actionable Insights}
\label{sec:actionable-insights}

Beyond providing an architecture for characterizing persistent agents, Pera offers a lens for examining what existing agent systems already realize and what remains underdeveloped for building persistent agents that provide continuing service in long-lived settings.
Organizing recent work through this lens reveals that many capabilities relevant to persistent agency have begun to emerge, but important gaps remain.
Together with the preceding case study, these gaps suggest several actionable directions for advancing persistent agents.

\paragraph{Adaptive sensing: deciding what, when, and how to sense.}

A persistent agent can be exposed to a continuous stream of signals from people, physical environments, digital systems, and the agent itself.
Recent work, including ContextAgent~\citep{yang2026contextagent} and ProAgent~\citep{yang2025proagent}, addresses parts of this problem through sensory context and on-demand perception~\citep{pu2025promemassist,lee2025sensible}.
Yet sensing over a long-lived setting raises a broader question: how can an agent continually decide what is worth perceiving as its setting and information needs change?
Adaptive sensing keeps the agent aware of relevant changes without processing every available signal continuously.

\begin{itemize}
    \item \textbf{Broad sensing scope.}
    Useful signals can take very different forms, such as a change in temperature, newly created files, or a change in a user's tone of voice.
    Therefore, a general lifecycle-perception interface should support signals from human, physical, digital, and internal sources rather than assume a single language or software interface~\citep{yang2025socialmind,zhang2026sensorlm}.
    Broadening the sensing interface can expose more signals relevant to future service and extend language agents beyond purely digital environments into the physical world.

    \item \textbf{Selective sensing.}
    Signals in a long-lived setting can be abundant, noisy, and unevenly useful, making continuous processing of everything both unnecessary and computationally expensive.
    A promising direction is to adapt which sources are monitored, when they are inspected, and at what frequency and granularity.
    For instance, routine periods require only lightweight monitoring, while an emerging anomaly can trigger finer-grained or more expensive sensing.
    Such selective allocation can make continuing perception more scalable while reducing unnecessary computation.

    \item \textbf{Evolving sensors.}
    Beyond selecting among available sensing sources, a persistent agent should also evolve how those sources are sensed as the setting changes~\citep{li2026act}.
    As the environment and relevant signals change over time, the agent should revise its software sensors, add new detectors or data sources, and use grounding actions to recalibrate, repair, replace, or introduce physical sensors.
    This makes the agent's sensing capability itself evolvable over its lifetime.
\end{itemize}

\paragraph{Signal representation: learning how to describe change.} 

Sensing exposes events, but lifecycle-level reasoning needs a representation in which perceived changes can be compared and related over time. 
Natural language is flexible, yet similar conditions can be described in very different ways, making long-term comparison and aggregation difficult~\citep{liu2023we,jia2025automated}. 
A promising direction is to map sensed events into more structured representations that need not themselves be natural language and can evolve with the setting. 

\begin{itemize}

\item \textbf{Domain-specific signal languages.}
Consider an agent managing a training cluster.
During one run, several workers slow down, communication latency rises, and job throughput drops.
Although these events together indicate the same underlying condition, such as a communication bottleneck, natural-language descriptions can vary substantially across runs.
Therefore, recognizing a similar condition weeks later requires more than matching surface descriptions.
A promising approach is to map sensed events into a domain-specific signal language with shared concepts and relations.
Such representations can provide a more consistent basis for comparing signals and reasoning over them across sources and over time.

\item \textbf{Cross-signal aggregation.}
Some changes become visible only when multiple signals are considered together.
A single execution failure does not establish a persistent problem, whereas repeated failures across tasks provide evidence that a previously effective procedure has become obsolete.
Therefore, a promising direction is to develop mechanisms that relate evidence across events and determine when multiple signals together indicate a higher-level condition~\citep{etzion2010event}.
This is especially important for weak signals that are individually insufficient but become informative when they recur over time.
Aggregating such evidence over long periods enables persistent agents to interpret persistent changes that cannot be understood from individual signals alone.
\end{itemize}

\paragraph{Context learning: thinking beyond reasoning over static model knowledge and curated context.}

Context learning is a key capability for persistent agents.
Over a long-lived setting, an agent accumulates large amounts of context that can be diverse, messy, and unfamiliar to the model, rather than a clean context prepared for a single task~\citep{dou2026cllife}.
Therefore, the challenge goes beyond reasoning over static model knowledge, as persistent agents need to continually learn from the context and use what they learn across future tasks.

\begin{itemize}

\item \textbf{Learning from real-world context.}
Real-world context rarely comes in a single, well-organized form.
For example, an employee's work context is scattered across conversations, documents, previous tasks, and application states, with useful information distributed across them rather than collected in one place~\citep{chen2026mmcl,wei2026clbench}.
A promising direction is to develop agents in environments where context is not cleanly packaged into a single document or conversation but instead accumulates across diverse sources and interactions over time.
Doing so can bring agents closer to understanding the same messy information environment in which people work and make decisions.

\item \textbf{Learning new knowledge from context.}
Much of the context encountered over an agent's lifetime contains knowledge that is new to the model, such as an unfamiliar workflow, a newly introduced policy, or a procedure learned through experience~\citep{si2026context}.
CL-bench~\citep{dou2026cl,dou2026cllife} begins to isolate this capability by requiring models to learn and apply knowledge, rules, and procedures absent from pre-training, while showing that substantial room for improvement remains.
Developing models that can reliably learn such new knowledge is a promising direction, enabling them to reason over unfamiliar knowledge from context rather than relying only on what is already learned in parameters.

\item \textbf{Learning the temporal validity of context.}
Knowledge learned from context does not remain valid indefinitely~\citep{huang2026survey}.
Its validity can change over time, for example, when an employee moves to a new project, a company policy is revised, or an earlier understanding is corrected by new evidence~\citep{sun2026preference}.
A promising direction is to make temporal validity part of context learning, so that agents learn not only what a piece of context means, but also when it applies and how later evidence changes its validity~\citep{su2026beyond}.
Recognizing such changes can further expose lifecycle-relevant signals that affect future service.

\end{itemize}

\paragraph{Proactivity: thinking beyond initiative within a given task.}

Recent agent research has increasingly focused on long-horizon task execution, where an agent is given a task and context and must solve a complex objective.
However, for persistent agents, an equally important challenge is determining when perceived changes and accumulated context warrant initiative, even without an explicitly specified task.
Proactivity can extend across different stages of persistent service, such as uncovering emerging needs from accumulated context, preparing or initiating useful work, and improving the agent itself when its current capabilities no longer fit future service.
Existing work has explored parts of this spectrum, such as proactively clarifying or acquiring information for an ongoing task~\citep{zhang2024ask} and, more recently, anticipating future needs from persistent context~\citep{lu2025proactive,pasternak2025beyond,xie2026pask}.

\begin{itemize}

\item \textbf{Discovering and anticipating needs from accumulated context.}
A promising direction is to enable persistent agents to proactively infer emerging needs from accumulated context~\citep{tang2026workspacebench10benchmarkingai,guo2026agoraarchivegroundedbenchmarkagentic,sun2026agenticdatabenchcomprehensivebenchmarkdata} and newly perceived signals, rather than waiting for explicit requests.
For example, an agent assisting an employee over months learns about their ongoing projects, deadlines, recurring information needs, and typical ways of working. A new document or a change in the project may reveal an opportunity for useful follow-up before the employee explicitly requests assistance~\citep{pasternak2025beyond,yang2026contextagent}.
Recent work~\citep{tang2026workspacebench10benchmarkingai,sun2026agenticdatabenchcomprehensivebenchmarkdata} has begun to explore such context-driven discovery, but it remains underexplored in long-lived settings, where some needs become apparent only across repeated interactions and tasks.
Combined with increasingly capable long-horizon execution, further progress could move agents from solving given tasks toward autonomously identifying and carrying out what needs to be done.

\item \textbf{Deciding when and how to act on anticipated needs.}
Identifying a possible need does not determine whether the agent should act.
For persistent agents that repeatedly take initiative, acting too aggressively can create interruptions or risks, while being overly conservative can make anticipation ineffective.
A promising direction is to calibrate proactive action based on factors such as confidence, expected benefit, and risk, with responses ranging from continued observation or background preparation to asking the user or carrying out low-risk actions \cite{horvitz1999principles}.
User feedback on previous initiatives can further help the agent learn which proactive behaviors are useful for a particular user and setting.
When an anticipated need instead reveals a capability gap, proactive preparation can include maintaining or improving the agent before the limitation affects future service.

\end{itemize}

\paragraph{Lifecycle control: coordinating work beyond a single task loop.}

Most agent control focuses on deciding what to do next within a given task.
Persistent agents additionally need to coordinate work across tasks, as episodic and lifecycle tasks can differ in urgency, depend on one another, compete for resources, and involve modifications to persistent components reused by future tasks~\citep{wei2025agent}.
This motivates lifecycle-level control over how work is coordinated and governed over time, including how tasks are prioritized and how persistent changes are specified and accepted.

\begin{itemize}

\item \textbf{Lifecycle-aware dispatch.}
Dispatching work in a long-lived setting differs from ordinary resource scheduling because tasks can have different implications.
Episodic and lifecycle tasks differ in urgency and long-term impact: user-facing work often carries immediate response requirements, whereas lifecycle work can determine the quality of many future tasks.
A promising direction is to make dispatch sensitive to the service implications of each task, such as its urgency, dependencies, and long-term impact.
Better lifecycle-aware dispatch could balance immediate responsiveness with long-term service quality, ensuring that important maintenance and adaptation are not indefinitely postponed without unnecessarily delaying urgent user-facing work.

\item \textbf{Specifying acceptable lifecycle changes.}
Because lifecycle tasks can modify components reused by future executions, an objective alone is insufficient to specify an acceptable change.
A promising direction is for the agent itself to specify, as part of the lifecycle task package, not only what should be achieved but also the conditions under which the resulting change can be accepted, such as its allowed scope, required evidence, and recovery requirements~\citep{chen2026cordon}.
This gives lifecycle control and review a clearer basis for deciding whether an update should be accepted or further revised~\citep{santos2026temporary}.
Reusable specifications for common forms of lifecycle change could make persistent adaptation more reliable across tasks~\citep{zhang2026you}.
\end{itemize}

\paragraph{Agent self-improvement: model and harness.}

Agent self-improvement concerns how accumulated context is translated into persistent changes to the agent itself.
Recent work increasingly uses \emph{agent harness} to refer to the non-model parts within an agent~\citep{huang2026memoharness,pan2026natural}.
Recent work has begun to enable agents to improve parts of their own harnesses, such as skills, tools, and memory mechanisms~\citep{chen2026failed,lin2026agentic,cai2026moss,lee2026meta}.
In Pera, the persistent agent core helps the agent proactively identify when change is needed.
Such improvement can target both the harness and the model.
The former provides an explicit and flexible space for rapid adaptation, while the latter can internalize repeatedly useful improvements into its parameters.
Despite their different forms, both can be improved along similar dimensions, including long-lived knowledge, decision-making procedures, and capabilities such as reasoning and learning.

\begin{itemize}

\item \textbf{Improving the improvement process.}
Self-improvement itself can become a capability that improves through accumulated context~\citep{chen2026failed,lin2026agentic}.
Evidence accumulated across tasks and over time can be used not only to change how the agent operates, but also to refine how it identifies and responds to the need for change.
The adaptive sensing discussed earlier provides one example: improving what, when, and how the agent senses alters the evidence available to lifecycle control.
Accumulated experience can likewise refine how lifecycle control responds, such as when to take initiative, which changes warrant intervention, and how much autonomy to exercise.
Making these mechanisms themselves evolvable could help the agent become better at identifying when and how to improve itself.
Compared with the evolution of individual harness components, this meta-level remains underexplored for persistent agents.

\item \textbf{Coordinated evolution.}
Existing self-improvement methods often target particular parts of the agent harness, but improvements to different components can interact in nontrivial ways~\citep{zhang2026self,cai2026moss,lee2026meta}.
For example, evolving a sensor exposes new signals that existing lifecycle control cannot yet use effectively, while introducing a new tool requires corresponding changes to the decision procedure that invokes it.
Therefore, a promising direction is to move from isolated component evolution toward coordinated evolution across the agent.
Such coordination can help improvements discovered in one component translate into better overall behavior rather than creating new mismatches elsewhere in the system.

\item \textbf{Safe self-improvement.}
Recent work has begun to validate self-generated agent modifications using execution feedback or regression testing~\citep{chen2026failed,zhang2026self}.
For persistent agents, the stakes are higher because an accepted modification can persist and affect many future tasks.
So calibrating the degree of autonomous modification according to the risk and reversibility of the change is a promising research direction.
Reversible changes can be explored more freely, for example through version-controlled updates (e.g., Git) that can be tested and rolled back, whereas more consequential modifications can require broader regression testing or staged acceptance, enabling safer and more autonomous self-improvement over time.

\item \textbf{Model self-improvement.}
The harness provides a flexible space for agent self-improvement, where new knowledge and decision procedures can be introduced and refined directly.
A further opportunity is to use such accumulated improvements to evolve the model itself.
Early work on agent fine-tuning has begun to transfer behaviors learned from agent trajectories into model parameters~\citep{lu2026skill0,zeng2024agenttuning}, but the broader connection between harness evolution and model learning remains underexplored.
A promising direction is for agents to self-improve their harnesses to discover and validate better decision procedures and learning and reasoning strategies, and then internalize repeatedly useful improvements into the model.
In this way, persistent experience could improve not only what the model knows, but also how it reasons and learns.

\end{itemize}

\paragraph{Agent evaluation: thinking beyond long-horizon tasks.}

Most agent evaluations still center on individual tasks, increasingly including tasks with longer and more complex execution (i.e., long-horizon tasks)~\citep{starace2025paperbench,kwa2026measuring,xu2026theagentcompany}.
Some work~\citep{zheng2025lifelongagentbench,dou2025evalearn,wang2026contextweaverealworldworkflowbenchmark} begins to move beyond isolated tasks by evaluating skill acquisition and transfer across interdependent task sequences.
Looking ahead, we expect agent evaluation to develop along several dimensions: toward tasks with greater real-world value, toward long-horizon execution within a task, and toward continuing service across the lifetime of a long-lived setting.

However, existing benchmarks still do not capture the full trajectory of long-lived service, where goals evolve over time, context continually accumulates and changes, and the agent must proactively support diverse tasks and adapt itself over time.
For example, in a knowledge-intensive professional setting, an agent assistant should support staff in their work over extended periods by working with accumulated and evolving context, supporting changing projects and needs, proactively identifying useful work, and adapting itself as the setting changes.
Such evaluation shifts the focus from \emph{how long an agent can execute a task} to how well it can sustain and improve service throughout extended real-world use.
Realistic and scalable benchmarks for this form of persistent service remain underdeveloped.

\section{Discussion}
\label{sec:dis}

In this section, we discuss two lines of work closely related to our conceptual framework.

\textbf{Continual learning and lifelong learning.}
We first consider two learning paradigms concerned with the continuous adaptation of models and agents.
Continual learning~\citep{parisi2019continual,de2021continual,shi2025continual} studies how models acquire knowledge from evolving data or task streams while retaining previously learned capabilities~\citep{zheng2025towards}.
Much of the work focuses on updating model parameters while mitigating catastrophic forgetting~\citep{mccloskey1989catastrophic,kirkpatrick2017overcoming,rebuffi2017icarl}.
We distinguish it from lifelong learning, which adopts the broader goal of accumulating reusable knowledge and experience across a lifetime~\citep{zheng2025lifelongagentbench,cai2025building}.
In the context of language models, prior work has studied how evolving knowledge and past experience can be incorporated into model parameters or external knowledge stores~\citep{lewis2020retrieval,borgeaud2022improving,meng2022locating,meng2022mass}.
Recent work has extended this perspective to language agents, examining how their memory, skills, and actions continually adapt through interaction with changing environments~\citep{zhao2024expel,cao2026remember,zhang2026memskill,yang2026evotool}.

However, these works leave several gaps.
First, existing formulations~\citep{yang2026autoskill,yang2026skillopt} center on streams of tasks, where experience from completed tasks is retained and reused to improve performance on subsequent tasks.
Yet realizing persistent agents requires a broader perceptual scope, in which agents continually and proactively perceive service-relevant signals from both external and internal sources, including changes that do not arrive as explicit tasks~\citep{deng2024towards,lu2025proactive,yang2026contextagent}.
Moreover, persistent agents must also operate prospectively, continually uncovering new contextual information as the setting evolves and using it to adapt their ongoing assistance.
Finally, existing work on persistent agents follows two main lines: either proposing specific lifelong learning approaches centered on learning from past experience, or surveying the broader research landscape by organizing existing work into taxonomies of agent capabilities and components~\citep{gao2025survey,cai2025building}.
Neither line, however, provides a conceptual architecture that both characterizes the operating principles of persistent agents and offers a unified lens for understanding how existing efforts contribute to them.
Pera addresses these gaps by proposing a perception-centered conceptual architecture and using it to retrospectively organize recent work.

\textbf{Frameworks for agents.}
A long line of work~\citep{maes1994agents,wooldridge1995intelligent,masterman2024landscape} has proposed general architectures and conceptual frameworks for organizing intelligent agents.
Early work in symbolic AI developed cognitive architectures such as Soar~\citep{laird1987soar}, which integrates perception, memory, learning, decision-making, and action within a unified cognitive system. 
The Common Model of Cognition~\citep{laird2017standard} later summarized the components and processing cycles shared by Soar and other influential cognitive architectures. 
Intentional architectures such as BDI~\citep{rao1995bdi,de2020bdi} instead organize practical reasoning around beliefs, desires, intentions, and plans.

With the rise of large language models, architectural thinking has been extended to language agents in multiple directions~\citep{wang2024survey,xi2025rise}.
CoALA~\citep{sumers2023cognitive} brings the cognitive-architecture perspective to language agents by structuring them around memory, internal and external actions, and decision procedures, and uses this structure to categorize existing agent methods.
Recent work has also explored the system infrastructure for language agents.
AIOS~\citep{ge2023llm,mei2024aios} organizes its runtime through an operating-system-like architecture that separates agent applications from the kernel responsible for execution and resource management.
Collectively, these language-agent frameworks organize particular aspects of agents, but do not provide a lifecycle-level account of how an agent continually perceives service-relevant changes and sustains evolving service responsibilities in a long-lived setting.

Another line of work places perception and adaptive control at the center of agent architectures.
The subsumption architecture~\citep{brooks1986robust,brooks1991intelligence} tightly couples sensory inputs with layered behaviors, enabling embodied agents to respond continuously to changes in their environment.
Hayes-Roth's architecture for adaptive intelligent systems (AIS)~\citep{hayes1995architecture} further supports the dynamic adaptation of perceptual strategies, reasoning tasks and methods, and control plans as operating conditions change.
The AIS architecture directly inspired Pera's emphasis on perception and control.

However, in these architectures, perception primarily serves to regulate the agent's behavior in support of current tasks.
For persistent agents, perception serves a broader purpose, including identifying emerging needs and determining how the agent's service should evolve with the setting~\citep{zheng2026lifelong}.
This difference in purpose leads to distinct designs for perception and control, and ultimately to different agent architectures, one oriented toward adapting current operation and the other toward sustaining and evolving service throughout the lifetime of a setting.
Moreover, these works predate the emergence of language models and do not explain how their underlying design principles should be reinterpreted and extended for language agents.
Pera builds on this tradition by making persistent and proactive assistance in long-lived settings a central architectural concern. 
It characterizes how external and internal signals give rise to lifecycle tasks, how lifecycle and episodic tasks interact, and how their interaction sustains and adapts the agent's service over time.

\section{Conclusion}
\label{sec:conclusion}

In this paper, we introduced Pera, a perception-centered architecture for persistent agents.
Pera extends existing cognitive frameworks for language agents by centering continual perception and lifecycle control, enabling adaptation across changing, long-lived settings.
We also used Pera to systematically organize a broad body of recent work and to present actionable insights for future research.
We hope Pera can provide a conceptual and architectural foundation for agents in the large, supporting the development of proactive, long-lived, and adaptive intelligent systems.


\bibliography{tmlr}
\bibliographystyle{tmlr}

\end{document}